\documentclass[conference]{IEEEtran}
\IEEEoverridecommandlockouts

\usepackage{cite}
\usepackage{amsmath,amssymb,amsfonts}
\usepackage{algpseudocode}
\usepackage{graphicx}
\usepackage{textcomp}
\usepackage{xcolor}
\usepackage{listings}
\usepackage{tabularx}
\usepackage{booktabs}
\usepackage{multirow}
\usepackage{mathpartir}
\usepackage{float}
\usepackage{placeins}
\usepackage[hidelinks]{hyperref}
\usepackage{pifont}                       % for \ding{182} etc.

\newcommand{\kw}[1]{\text{\ttfamily #1}}
\newcommand{\akw}[1]{\text{\ttfamily\symbol{92}#1}}

\def\BibTeX{{\rm B\kern-.05em{\sc i\kern-.025em b}\kern-.08em
    T\kern-.1667em\lower.7ex\hbox{E}\kern-.125emX}}

\begin{document}

% ============================================================================
% Title and author block (IEEEtran style)
% ============================================================================
\title{
%Beyond Vacuous Frames:\\
       AutoFraming: Synthesizing  
       Dynamic Frames of C Code with LLMs %
}

% \author{
% \IEEEauthorblockN{Anonymous Authors}
% }
% \textit{Department of \ldots} \\
\author{
\IEEEauthorblockN{Han Zhou}
\IEEEauthorblockA{
\textit{University of Missouri--Kansas City} \\
Kansas City, USA \\
hzb4f@umkc.edu
}
\and
\IEEEauthorblockN{Yu Luo}
\IEEEauthorblockA{
\textit{University of Central Missouri} \\
Warrensburg, USA \\
yuluo@ucmo.edu
}
\and
\IEEEauthorblockN{Dianxiang Xu}
\IEEEauthorblockA{
\textit{University of Missouri--Kansas City} \\
Kansas City, USA \\
dxu@umkc.edu
}
}

\maketitle

% ============================================================================
% Abstract and keywords (IEEEtran style)
% ============================================================================
\begin{abstract}

Large language models (LLMs) have shown considerable promise in synthesizing formal specifications from source code. However, existing approaches primarily focus on functional contracts and loop invariants, while largely neglecting frame conditions, which specify the memory locations that may or may not be modified during execution and constitute an essential part of formal specifications.
To address this limitation, we propose an LLM-based approach for synthesizing ACSL frame specifications alongside functional contracts from C code. Our approach analyzes code property graphs (CPGs) to extract frame-relevant information from the input code and leverages this information to guide the LLM in generating accurate frame conditions. The resulting specifications capture both region-separation constraints and data dependencies that govern variable modifications.
Experimental results demonstrate the effectiveness of our approach. On a benchmark of 84 ACSL-compliant functions from non-trivial real-world systems, our method successfully synthesizes error-free frame specifications for 92.9\% of the functions, representing a 179\% relative improvement over state-of-the-art techniques. On a second benchmark consisting of 51 functions, the success rate increases from 80.4\% to 96.1\%. Furthermore, when integrated as an auxiliary component into existing LLM-based ACSL-synthesis frameworks, our approach improves frame-specification generation performance by up to 61.9\%.

\end{abstract}


% ============================================================================
\section{Introduction}
\label{sec:intro}

Large language models (LLMs) have recently shown strong potential for generating formal specifications from source code. For C programs, several approaches~\cite{George2025Seeking,George2025Specify,Christian2023,Cheng2024, wang2025} have explored using LLMs to synthesize ACSL specifications, often with feedback from static analyzers or verification tools. These techniques can generate useful functional specifications, such as preconditions, postconditions, and loop invariants, and can substantially reduce the manual effort required for deductive verification. However, little attention has been paid to frame conditions: the part of a specification that describes what memory a function may affect, what values those effects depend on, and what memory must remain independent. Without accurate frame information, a verified function cannot be safely reused by callers, because the caller cannot know which parts of the program state are preserved.

This limitation is especially important for C programs. Recent approaches such as AutoSpec~\cite{Cheng2024} and Preguss~\cite{wang2025} push LLM-based specification generation further by combining models with program information and verification feedback. However, their treatment of frame information remains limited. They primarily optimize for whether the generated specification can help discharge verification conditions, while frame completeness is not directly measured. In particular, evaluations based mainly on WP proof success may overlook missing memory-dependence information or imprecise aliasing assumptions, because a proof can succeed even when the contract does not fully describe what state a caller may rely on after the function returns. This gap is difficult to close for C: pointers, arrays, pointer arithmetic, loops, and library calls can hide which memory regions may overlap and which input values may influence each write. As a result, a generated contract may look correct to the verifier but still be too weak for precise modular reasoning. Section~\ref{sec:background} shows a small example where an apparently verified contract still permits an invalid aliased call.

To address this problem, we propose AutoFraming, an approach for synthesizing frame specifications for C programs. AutoFraming analyzes frame-relevant information from a code property graph (CPG) \cite{yamaguchi2014modeling}, including possible memory interference, caller-visible writes, loop-indexed write ranges, and data dependencies. It then projects this information into a fixed-frame skeleton that captures which memory regions must be kept independent and which inputs may influence each written location. The LLM is used to assemble the full contract and repair remaining verification failures, while the frame-critical structure is generated and preserved during refinement.

We evaluate AutoFraming on two benchmarks: 84 functions from real-world C systems and 51 academic verification examples. On the real-world benchmark, AutoFraming synthesizes error-free frame specifications for 92.9\% of functions, a 179\% relative improvement over the state-of-the-art baseline. On the academic benchmark, it improves the error-free frame rate from 80.4\% to 96.1\%. When used as an auxiliary frame-synthesis component for existing specification-generation tools, AutoFraming further improves error-free frame performance by up to 61.9\%. Ablation results show that the CPG-derived frame information and deterministic projections are essential to the improvement.

The contribution of this paper is summarized as follows: AutoFraming provides an effective approach to the novel problem of synthesizing
error-free frame conditions with LLMs.
It conducts frame analysis, including caller-visible roots, access windows, interfering pointer pairs, write-set locations, and dependence footprints. It then projects frame-related information into ACSL specification, including region separation for caller-visible memory regions and data dependence for write-set locations. 
Moreover, AutoFraming can be used in other ACSL synthesis frameworks, such as Preguss and AutoSpec, to improve their capability of synthesizing frame conditions. 

\section{A Motivating Example}
\label{sec:background}

Consider the C function \texttt{parse\_value}, shown below, which reads two values from the memory region pointed to by \texttt{buf} and stores their sum in the memory location pointed to by \texttt{eaten}.  
 
\begin{lstlisting}[language=C,
  caption={Source code of the motivating example.},
  label={lst:parse-value}]
int parse_value(const uint16_t *buf, uint16_t len, uint16_t *eaten) {
    if (len < 2 || buf == NULL || eaten == NULL) return -1;
    *eaten = buf[0];
    *eaten = (uint16_t)(*eaten + buf[1]);   
    return 0;
}
\end{lstlisting}

To call \texttt{parse\_value}, the client needs to provide actual parameters for  \texttt{*buf} and \texttt{*eaten}, such as the following:  
 
\begin{lstlisting}[language=C]
uint16_t buf[2] = { 3, 5 };
uint16_t out;
parse_value(buf, 2, &out);      
\end{lstlisting}
 
\noindent

Without a formal specification for \texttt{parse\_value}, clients have no precise way to determine whether a call satisfies the function's assumptions or achieves its intended behavior. To address this issue, we can prompt an LLM to synthesize the following ACSL~\cite{baudin2008acsl} specification directly from the source code of \texttt{parse\_value}:
 
\begin{lstlisting}[language=C,
  caption={Preliminary ACSL  specification generated by prompting an LLM}, label={lst:naive}]
/*@ requires len > 0;
    requires buf != \null;
    ensures \result == 0 || \result == -1; */
int parse_value(const uint16_t *buf, uint16_t len, uint16_t *eaten);
\end{lstlisting}
 
\noindent
The generated comment enclosed by \texttt{/*@} and \texttt{*/} defines the ACSL contract for \texttt{parse\_value}, including its precondition (the \texttt{requires} clauses) and postcondition (the \texttt{ensures} clause). The precondition specifies the conditions that callers must satisfy before invoking \texttt{parse\_value}; specifically, \texttt{len} must be positive and \texttt{buf} must be non-null. The postcondition specifies the property that must hold when \texttt{parse\_value} returns, namely that the return value is either \texttt{0} or \texttt{-1}.

The ACSL specification can be verified with Frama-C and one of its  plug-ins (e.g., WP or Eva) ~\cite{cuoq2012framac}. Given a function and its ACSL contract, Frama-C generates a set of proof obligations, called \emph{goals}, each expressed as a logical formula whose validity establishes a specific aspect of the contract. These goals are then dispatched to automated SMT solvers. The proportion of discharged goals, referred to as the \emph{goal ratio} (proved/total), indicates the extent to which the contract has been verified. A goal ratio of $1$ (i.e., a full proof) means that every proof obligation has been discharged, establishing that the implementation satisfies the specified contract for all executions. Otherwise, the result is only a partial proof, indicating that some proof obligations remain unresolved. It is important to note, however, that a full proof does not necessarily imply that the code is correct. Rather, it only guarantees conformance to the given specification. If the specification is incomplete or inaccurate, the implementation may still satisfy it while failing to capture the intended behavior. For example, Frama-C with the WP plugin reports a full proof for the preliminary LLM-generated ACSL contract shown above, even though the contract only partially captures the intended behavior of \texttt{parse\_value}. The generated precondition is overly restrictive. It requires \texttt{len} to be positive and \texttt{buf} to be non-null, even though \texttt{parse\_value} handles the cases where \texttt{len} is non-positive or \texttt{buf} is \texttt{NULL} without error.

\noindent
To improve the quality of  synthesized ACSL specifications, existing approaches leverage the results of static code analysis to prompt LLMs and refine the specifications. 
The following ACSL specification is generated by Preguss ~\cite{wang2025}, the state-of-the-art method:

\begin{lstlisting}[language=C,
  caption={ACSL specification generated by existing approaches}, label={lst:existing}]
/*@ requires \valid_read(buf + (0..len-1));
    requires \valid(eaten);
    assigns  *eaten;
    ensures  \result == 0 || \result == -1; */
int parse_value(const uint16_t *buf, uint16_t len, uint16_t *eaten);
\end{lstlisting}
 
\noindent 
The \texttt{requires} clauses specify the validity conditions for \texttt{buf} and \texttt{eaten}. The \texttt{assigns} clause, \texttt{assigns *eaten;}, defines the function's frame condition by specifying that only the memory location referenced by \texttt{eaten} may be modified during execution. Consequently, all other program state, including \texttt{buf}, \texttt{len}, and the memory referenced by \texttt{buf}, must remain unchanged. More generally, the \texttt{assigns} clause consists of a list of variable names or memory locations (referred to as a write set or footprint).  

However, the above frame specification is insufficient even though the verification result is a full proof. Consider the following call:

\begin{lstlisting}[language=C]
uint16_t buf[2] = { 3, 5 };
parse_value(buf, 2, &buf[1]);   
\end{lstlisting}

\noindent
The call satisfies 
both precondition clauses, i.e., \texttt{buf} is readable and \texttt{\&buf[1]} is
valid. Although the intended value for \texttt{*eaten} (i.e., \texttt{buf[1]}) is
\texttt{buf[0] + buf[1] = 8}, the actual outcome is \texttt{3 + 3 = 6} because \texttt{parse\_value} reads \texttt{buf[1]}
\emph{after} writing through \texttt{*eaten}, so the write clobbers
\texttt{buf[1]} first and the call computes \texttt{3 + 3}. 

The problem arises because the same object, \texttt{buf}, is passed to both pointer parameters, \texttt{*buf} and \texttt{*eaten}. However, \texttt{parse\_value} operates correctly only when these pointers refer to separate, non-overlapping memory locations. Therefore, the ACSL specification must explicitly capture this separation requirement to prevent invalid function calls and ensure correct verification.

In summary, existing LLM-based approaches to ACSL specification synthesis do not explicitly address the frame problem, particularly dynamic frame conditions such as the memory separation requirement illustrated above. Dynamic frame conditions precisely specify the memory locations that a function is permitted to read or modify, ensuring that changes to one object cannot inadvertently affect the state of another unrelated object. The goal of this paper is to enable LLMs to synthesize ACSL specifications with accurate frame conditions for C programs that are ACSL-compliant, i.e., programs whose behavior can be completely specified in ACSL. The following specification for the example illustrates the desired result:

\begin{lstlisting}[language=C,
  caption={ACSL specification with dynamic frames}, label={lst:ours}]
/*@ requires \valid_read(buf + (0..len-1));
    requires \valid(eaten);
    requires \separated(buf + (0..len-1), eaten);  
    assigns  *eaten                                 
             \from buf[0..len-1], len;          
    ensures  \result == 0 || \result == -1; */
int parse_value(const uint16_t *buf, uint16_t len, uint16_t *eaten);
\end{lstlisting}

The additional \texttt{requires} clause,
\texttt{\textbackslash separated(buf + (0..len-1), eaten);}
specifies that the memory region referenced by \texttt{buf} must be disjoint from the memory location referenced by \texttt{eaten}. The \texttt{assigns} clause,
\texttt{*eaten \textbackslash from buf[0..len-1], len;}
defines both the function's write set and the data dependencies of the modified location, stating that only \texttt{*eaten} may be updated and that its value depends solely on \texttt{buf[0..len-1]} and \texttt{len}.
Synthesizing the region separation requirement and the data dependency of the write set is a challenging problem. Because C programs do not expose their memory footprint directly in the source syntax. Pointer aliasing, array ranges, pointer arithmetic, loops, globals, and library calls can hide both the regions that may interfere with one another and the inputs that may affect each write. Therefore, synthesizing accurate dynamic frames requires recovering caller-visible memory regions, separation assumptions, and write dependencies from program structure rather than treating frame clauses as ordinary text to be generated.

\section{The design of AutoFraming}
\subsection{Overview}
\label{sec:framework}

\begin{figure}[!t]
  \centering
  \includegraphics[width=\linewidth]{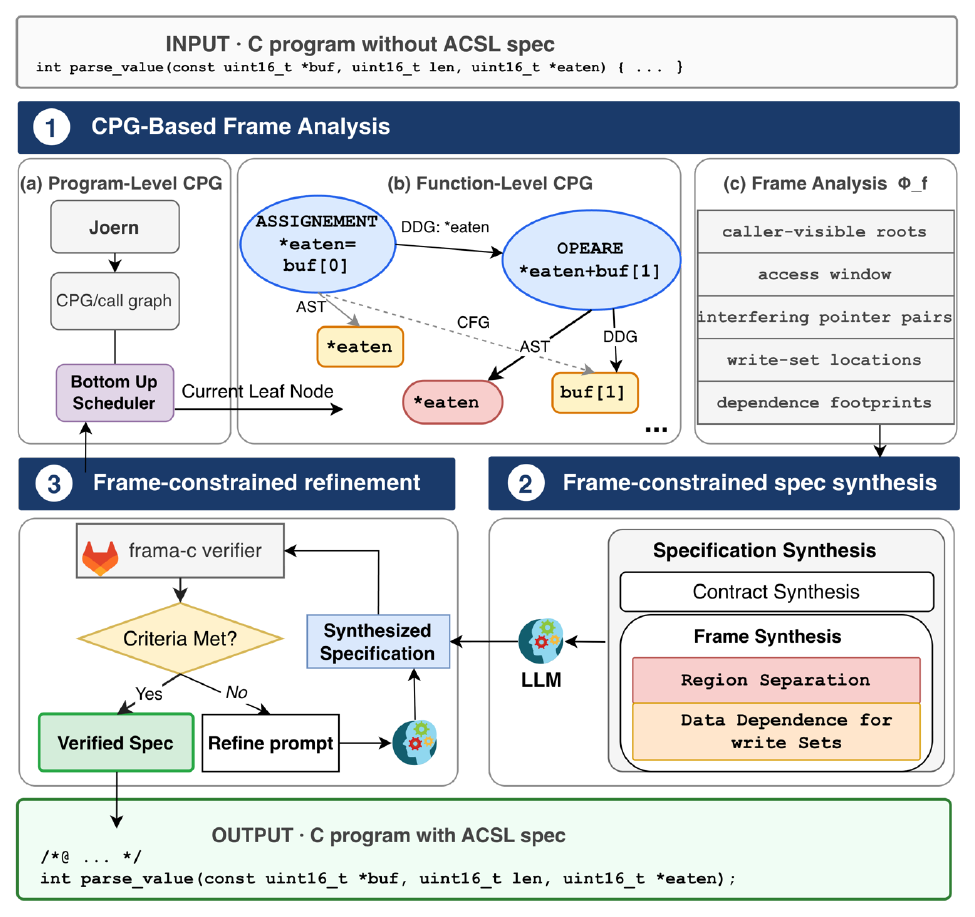}
  \caption{AutoFraming Pipeline for CPG-guided Frame Specification Synthesis.}
  \label{fig:framework}
\end{figure}

AutoFraming leverages the code property graph (CPG) representation of C code  to synthesize its frame specifications. Given an unannotated function, AutoFraming generates an ACSL specification whose frame part describes not only the memory locations that may be modified, but also the memory regions that must remain disjoint and the caller-visible inputs that may influence each write. Although AutoFraming outputs a complete ACSL specification, this paper focuses on the synthesis of frame conditions. Figure~\ref{fig:framework} shows the three-stage workflow of AutoFraming. 

Stage 1: AutoFraming first constructs the call graph and analyzes functions in bottom-up order, ensuring that verified callee contracts are available when analyzing their callers. For each function $f$, AutoFraming builds a CPG and extracts the frame-related information $\Phi_f$, which captures the caller-visible roots of memory accesses, accessed regions of pointer roots, pointer pairs whose aliasing may interfere with caller-visible writes, write-set locations used as frame carriers, and the dependence footprint of each write-set location.

Stage 2: AutoFraming projects $\Phi_f$ into a frame skeleton. The separation projection turns interfering pointer regions into disjointness requirements, while the dependence projection attaches caller-visible dependence footprints to write-set locations. AutoFraming treats projected frame obligations as required constraints.
A candidate is rejected if it deletes or weakens them, but the current
implementation may still allow equivalent or stronger rewritings that
change the proof burden.

Stage 3: AutoFraming asks the LLM to assemble a verifier-checkable contract around the frame skeleton and uses Frama-C feedback to repair remaining failures. During refinement, candidate contracts are accepted only if they preserve the projected frame conditions, avoid degenerate frames, and make proof progress. This design lets AutoFraming use the LLM for auxiliary annotations and repair while keeping the frame-critical content grounded in program analysis.

\subsection{CPG-Based Frame Analysis}
\label{sec:cpg}

For each function $f$, AutoFraming extracts from its CPG the information needed to synthesize the dynamic frame. A CPG overlays three program views on one graph~\cite{yamaguchi2014cpg}: the abstract syntax tree (AST), which exposes declarations and assignment targets; the
control-flow graph (CFG), which exposes branches, loops, and execution paths; and the program dependence graph (PDG), which exposes data and control dependencies. This combination is essential because frame synthesis is not a purely syntactic task: write-set locations come from syntax, region extents often come from loops and guards, and dependence footprints require following value and control flow.

AutoFraming represents the frame-related information as
\[
\Phi_f =
(\mathcal{R}_f,\Omega_f,\mathcal{I}_f,\mathcal{L}_f,\Delta_f).
\]
The first three components help address the region-separation issue:
which caller-visible regions must be disjoint? The last two deal with the data dependency issue: what does each write set location depend on?

\paragraph{CPG-to-information extraction.}
AutoFraming derives each component of $\Phi_f$ by querying specific CPG
nodes and edges rather than by passing the whole graph to the LLM. AST
nodes provide syntactic roots and memory-access forms: parameter and
global declarations identify caller-visible roots; dereference,
array-subscript, field-access, assignment, and update nodes identify
reads and writes; and declaration and cast nodes provide type information
for alias compatibility. CFG edges provide the execution context of these
accesses: loop headers, back edges, and branch guards are used to recover
loop-indexed access ranges and conditional bounds. PDG edges provide
dependence information: data-dependence edges connect values to the
expressions they influence, while control-dependence edges connect guards
to the writes whose execution they control. Together, these CPG views
determine the five information components used by AutoFraming:
caller-visible roots, access windows, interfering pointer pairs,
write-set locations, and dependence footprints.

\begin{table}[t]
\centering
\footnotesize
\setlength{\tabcolsep}{4pt}
\renewcommand{\arraystretch}{1.2}
\caption{The frame-information set $\Phi_f$ extracted from the CPG.}
\label{tab:notation}
\begin{tabularx}{\columnwidth}{@{}l X l c@{}}
\toprule
\textbf{Component} & \textbf{Meaning} & \textbf{CPG view} & \textbf{Used by} \\
\midrule
$\mathcal{R}_f$
  & caller-visible roots: parameters and globals touched by $f$
  & AST & Sep, Dep \\
$\Omega_f$
  & access window of each pointer root
  & AST, CFG & Sep, Dep \\
$\mathcal{I}_f$
  & interfering pointer-root pairs requiring separation
  & AST, PDG & Sep \\
$\mathcal{L}_f$
  & normalized write-set locations used as \texttt{assigns} carriers
  & AST, CFG & Dep \\
$\Delta_f$
  & dependence footprint of each write set location
  & PDG & Dep \\
\bottomrule
\end{tabularx}
\end{table}

\paragraph{Caller-visible roots.}
$\mathcal{R}_f$ contains the formal parameters and globals that the
function may read or write, each recorded with its type. These roots are
the only memory roots that can appear in the external frame of $f$.
Local variables are excluded because their storage does not escape the
function call. Pointer roots are recorded with their pointed-to types,
since type compatibility determines whether two roots may denote
overlapping storage. 
For a pointer-typed root $p$ with C type $T *$, we call $T$ the
\emph{pointed-to type} of $p$. For example, the pointed-to type of
\texttt{const uint16\_t *buf} is \texttt{uint16\_t}, and the pointed-to
type of \texttt{uint16\_t *eaten} is also \texttt{uint16\_t}. Hence,
\texttt{buf} and \texttt{eaten} may refer to overlapping
\texttt{uint16\_t} storage at a call site.

AutoFraming uses pointed-to types as a conservative
alias-compatibility filter. Two pointer roots are considered compatible
when their accesses may refer to the same kind of object storage. This
includes roots with the same pointed-to type after removing qualifiers
such as \texttt{const} and \texttt{volatile}, roots accessed through
byte-level pointer types such as \texttt{char *} or
\texttt{unsigned char *}, roots accessed through generic
\texttt{void *} interfaces, and roots whose type information is weakened
by casts. Conversely, roots with clearly different non-byte pointed-to
types, such as \texttt{int *} and \texttt{double *}, are filtered out
when the function contains no casts, byte-level accesses, or generic
pointer operations that could make overlap relevant.

\paragraph{Access windows.}
For each pointer root $p \in \mathcal{R}_f$, $\Omega_f(p)$ records the
entry-expressible region of $p$ that the function may access:
\[
\Omega_f(p)
=
\mathsf{window}\bigl(\mathsf{Accesses}_f(p)\bigr).
\]
The access set $\mathsf{Accesses}_f(p)$ is collected by traversing AST
memory-access nodes rooted at $p$, including dereferences, array
subscripts, field accesses, and summarized library calls. It then uses
CFG loop and guard edges to attach path conditions and ranges to these
accesses. A constant access such as \texttt{p[c]} contributes offset
$c$; a loop-indexed access such as \texttt{p[i]} contributes the loop
range recovered from the loop header and back edge; and an access guarded
by a condition such as \texttt{i < n} contributes a range bounded by $n$
when $n$ is available at function entry. These
windows allow AutoFraming to emit separation constraints over accessed
regions rather than over pointer names alone.

\paragraph{Interfering pointer pairs.}
\[
\begin{aligned}
\mathcal{I}_f =
\{\,\{p,q\} \mid\;&
p,q \in \mathcal{R}_f
\land \mathsf{ptr}(p)
\land \mathsf{ptr}(q) \\
&{}\land \mathsf{compatible}(p,q)
\land \mathsf{interferes}_f(p,q)
\,\}.
\end{aligned}
\]
Here, $\mathsf{compatible}(p,q)$ applies the pointed-to-type filter:
it keeps same-type, byte-level, generic-pointer, and cast-related pairs,
and filters out clearly incompatible typed accesses. The predicate
$\mathsf{interferes}_f(p,q)$ checks whether aliasing the two roots can
change a caller-visible write or the value written:
\[
\begin{aligned}
\mathsf{interferes}_f(p,q) \triangleq\;&
\bigl(\mathsf{writes}_f(p)
      \land \mathsf{affectsWrite}_f(q)\bigr) \\
&{}\lor
\bigl(\mathsf{writes}_f(q)
      \land \mathsf{affectsWrite}_f(p)\bigr).
\end{aligned}
\]
A root satisfies $\mathsf{affectsWrite}_f$ when it either roots a write
or is read into a value that flows to a write. This interference test
prevents AutoFraming from over-restricting callers by separating
unrelated pointer parameters.

% \paragraph{Write sets.}
% $\mathcal{L}_f$ contains the caller-visible locations that may be
% modified by $f$, normalized into ACSL assignable locations. AutoFraming
% first collects assignment left-hand sides rooted in $\mathcal{R}_f$,
% including scalar roots, array cells, structure fields, and dereferences.
% When $w=A[i]$ is indexed by a loop variable and
% $\mathsf{range}_f(i)=(\mathit{lo},\mathit{hi})$, AutoFraming normalizes
% the target to the whole loop-written range:
% \[
% \mathsf{norm}_f(w)=
% \begin{cases}
% A[\mathit{lo}\,..\,\mathit{hi}-1],
%   & \text{if } w=A[i],\\
% w,
%   & \text{otherwise.}
% \end{cases}
% \]
% The write-set set is therefore
% \[
% \mathcal{L}_f =
% \{\, \mathsf{norm}_f(w)
% \mid w \in \mathsf{Writes}_f^{vis} \,\},
% \]
% where $\mathsf{Writes}_f^{vis}$ denotes the caller-visible write targets
% of $f$.
% These locations supply the left-hand side of ACSL \texttt{assigns}
% clauses. In AutoFraming, however, the write location is not the main
% synthesis target; it is the carrier to which dependence information is
% attached.

\paragraph{Write-set locations.}
$\mathcal{L}_f$ contains the caller-visible locations that may be
modified by $f$, normalized into ACSL assignable locations. AutoFraming
first collects write targets from AST write nodes in the CPG, including
assignment left-hand sides, array-cell updates, structure-field updates,
and dereferences. We use $w$ to denote one such write target. A target
$w$ is kept only when it can be expressed using a caller-visible root in
$\mathcal{R}_f$, such as a parameter or global. If a write is performed
through a local pointer alias, AutoFraming uses CPG data-dependence edges
only to rewrite the target back to its caller-visible root; the local
alias itself is never emitted in an ACSL frame clause.

When a write target $w$ is an array access indexed by a loop variable,
for example $w=A[i]$, AutoFraming uses CFG loop and guard edges to
recover $\mathsf{range}_f(i)=(lo,hi)$ and normalizes the target to the
whole loop-written range:
\[
\mathsf{norm}_f(w)=
\begin{cases}
A[lo..hi-1], & \text{if } w=A[i],\\
w, & \text{otherwise.}
\end{cases}
\]
The write-set location set is therefore
\[
\mathcal{L}_f =
\{\, \mathsf{norm}_f(w) \mid w \in \mathsf{Writes}^{vis}_f \,\},
\]
where $\mathsf{Writes}^{vis}_f$ denotes the collected write targets that
can be expressed using caller-visible roots. These locations supply the
left-hand side of ACSL \texttt{assigns} clauses. In AutoFraming, however,
the write location is not the main synthesis target; it is the carrier to
which dependence information is attached.
\paragraph{Dependence footprints.}
For each write set location $L \in \mathcal{L}_f$, $\Delta_f(L)$ contains
the caller-visible inputs whose values may influence the value written to
$L$:
\[
\Delta_f(L)
=
\{\, r \in \mathcal{R}_f
\mid
r \leadsto^{data/control}_f \mathsf{write}(L)
\,\}.
\]
The relation $\leadsto^{data/control}_f$ is computed from the PDG. It
includes data dependencies, where a root contributes to the expression
being written, and control dependencies, where a root controls whether a
write occurs. Local intermediate variables are followed transitively but
are not reported; the final dependence set contains only caller-visible
inputs. Pointer reads are lifted to accessed regions using $\Omega_f$, so
a loop that writes an output from \texttt{buf[i]} contributes the
accessed region of \texttt{buf}, rather than only the syntactic root
\texttt{buf}.

\subsection{Synthesis of Region-Separation Constraints}
\label{sec:region-separation}

The first dynamic-frame synthesis issue is region separation:
for pointer-accessed memory, a frame specification must identify the
caller-visible regions that must be disjoint for the declared frame to be
valid. If two such regions may overlap at a call site, a write through
one pointer can affect a read or write through another, making the
specified frame unsound or incomplete for modular reasoning.

Regional separation analysis is difficult for C code because the
required disjointness is not explicit in the source text. The same
function body can be called with separated or aliased actual arguments,
and the relevant regions may be hidden behind pointer arithmetic,
loop-indexed accesses, local pointer aliases, casts, or library calls.
Thus, the synthesis problem is not merely to emit a
\texttt{\textbackslash separated} clause, but to determine which
caller-visible pointer regions require such a clause and what region
extent should be used.

AutoFraming uses frame information extracted from the CPG to synthesize region-separation requirements. The interfering-pair
component $\mathcal{I}_f$ identifies caller-visible pointer roots whose
aliasing can affect a caller-visible write, while the access-window
component $\Omega_f$ identifies the region of each root that the function
may access. The separation projection then maps each pair in
$\mathcal{I}_f$ and its corresponding access windows to an explicit
caller-side \texttt{\textbackslash separated} precondition.

For each $\{p,q\}\in\mathcal{I}_f$, AutoFraming emits
$\mathsf{Sep}_f(p,q)$.
\[
\begin{aligned}
\rho_p &= \mathsf{region}(p,\Omega_f(p)),\\
\rho_q &= \mathsf{region}(q,\Omega_f(q)),\\
\mathsf{Sep}_f(p,q)
&= \kw{requires}~
   \akw{separated}(\rho_p,\rho_q)\kw{;}.
\end{aligned}
\]
The function $\mathsf{region}$ converts a pointer root and its access
window into an entry-expressible ACSL memory region, such as
\texttt{p}, \texttt{p+(0..n-1)}, or another slice expressible at function
entry. A constant offset contributes a singleton extent; a loop-indexed
offset contributes the loop range; and a guarded access contributes the
guard bound when that bound is available at function entry.

%This rule is precise other than separating all pointer parameters. 
Type compatibility alone is insufficient: many C functions take multiple
pointers of the same type without requiring them to be disjoint.
AutoFraming emits a separation constraint only when aliasing the two
roots can affect the function's caller-visible frame. This distinction is
important for C APIs such as copy, parse, and decode routines, where
source and destination pointers have compatible types but only some
pairs need disjointness.

When an access window cannot be expressed at function entry,
AutoFraming emits the strongest entry-expressible approximation available
and sends the candidate to the verification-repair stage. Such a
candidate is not accepted merely because the clause was generated; it
must still preserve the projected frame and discharge the relevant frame
obligations.

\subsection{Synthesis of Data Dependencies for Write Sets}
\label{sec:data-dependence}

The second frame synthesis issue is data dependence for
write sets: for each location that a function may modify, the frame
specification must identify the caller visible inputs that may influence
the value written there. An ACSL \texttt{assigns} clause without a
\texttt{\textbackslash from} footprint states where the function may
write, but not what information the written value may depend on. Such a
frame is incomplete for modular reasoning because callers cannot
determine which parts of the prestate are relevant to the modified
location.
Inferring this footprint is difficult in C because the dependence of a
written value is often separated from the final assignment. The value may
flow through local temporaries, pointer aliases, loop carried updates,
control predicates, globals, or summarized library calls before reaching
the write. As a result, reading the left hand side and right hand side of
the final assignment is not enough to recover the frame dependence.

AutoFraming handles this by using the write-set locations in
$\mathcal{L}_f$ as carriers and attaching to each one the dependence
footprint $\Delta_f(L)$ extracted from the CPG. The projection therefore
turns the internal value-flow information for each write into the ACSL
form expected by modular verification.

For each normalized write set location $L \in \mathcal{L}_f$,
the data dependence projection attaches $\Delta_f(L)$:
\[
\inferrule*[right={\textsc{Dep}}]
  {
    L \in \mathcal{L}_f \\
    \Delta_f(L)=\{r_1,\ldots,r_m\} \\
    m > 0
  }
  {
    \texttt{assigns}~L~
    \texttt{\textbackslash from}~r_1,\ldots,r_m\texttt{;}
  }
\]
If $\Delta_f(L)=\emptyset$, AutoFraming emits:
\[
\inferrule*[right={\textsc{Dep-Const}}]
  {
    L \in \mathcal{L}_f \\
    \Delta_f(L)=\emptyset
  }
  {
    \texttt{assigns}~L~
    \texttt{\textbackslash from}~
    \texttt{\textbackslash nothing}\texttt{;}
  }
\]

The dependence footprint includes both data and control dependencies. A
root is included in $\Delta_f(L)$ if its value flows into the expression
written to $L$, or if it controls whether a write to $L$ is executed. For
example, in a branch that writes different constants to \texttt{*out}
depending on \texttt{flag}, the output depends on \texttt{flag} even
though \texttt{flag} does not appear on the right-hand side of either
assignment. Capturing this control dependence is necessary for the
\texttt{\textbackslash from} clause to describe the value written, not
just the memory location modified.

The result of the two projections is a frame skeleton $S_f$. The
region separation projection contributes the \texttt{\textbackslash separated}
preconditions, and the data dependence projection contributes the
\texttt{assigns}\,\ldots\,\texttt{\textbackslash from} clauses. Thus,
AutoFraming's synthesis target is not merely the static write set, but
the dynamic-frame information attached to that write set.

\subsection{Frame-constrained contract synthesis}
\label{sec:contract-assembly}

The two projection rules produce a frame skeleton $S_f$ consisting of
\texttt{\textbackslash separated} preconditions and
\texttt{assigns}\,\ldots\,\texttt{\textbackslash from} clauses. This
skeleton is the output of the frame-constrained projection step. AutoFraming
does not ask the LLM to invent or reinterpret it. Instead, $S_f$ is
inserted into the prompt as a set of required
frame obligations in the candidate contract.

Frama-C verifier, however, checks frame clauses only inside a complete ACSL
contract. A frame skeleton can introduce proof obligations that require
additional surrounding annotations. For example, a separated region
$p+(0..n-1)$ requires preconditions establishing that the region is
well-formed and valid at function entry; an assigned slice
$A[lo..hi-1]$ may require loop annotations showing that the loop writes
only that range; and a \texttt{\textbackslash from} footprint may require
auxiliary assertions or invariants that expose how values flow through
local temporaries. AutoFraming therefore asks the LLM to synthesize a
contract envelope $E_f$ around the locked frame skeleton:
\[
C_f^{(0)} = E_f \cup S_f
\qquad\text{with}\qquad
S_f \subseteq C_f^{(0)} .
\]
The envelope may contain validity preconditions, simple functional
postconditions, loop invariants, loop variants, loop assigns clauses, and
auxiliary assertions needed by the verifier.

The frame information $\Phi_f$ guides this envelope synthesis. Access
windows $\Omega_f$ suggest validity requirements such as
\texttt{\textbackslash valid\_read} for input regions and
\texttt{\textbackslash valid} for written regions. write-set locations in
$\mathcal{L}_f$ guide loop assigns annotations and frame-preserving loop
invariants. Dependence footprints $\Delta_f(L)$ identify the inputs that
are relevant to postconditions over assigned outputs. When a function
calls another function, AutoFraming provides the verified callee contract
as context, so the LLM can use the callee's frame and functional clauses
rather than re-analyzing the callee body.

This design separates the role of program analysis from the role of the
LLM. Program analysis and frame-constrained projection synthesize the frame-critical
content $S_f$. The LLM supplies only the verifier-facing envelope $E_f$
needed to make $S_f$ checkable in a complete ACSL contract. As a result,
AutoFraming avoids relying on the LLM to guess dynamic-frame clauses from
source code alone, while still using the LLM to fill
in auxiliary annotations and functional context that are necessary for verifier to discharge the generated frame.

\subsection{Frame-constrained refinement}
\label{sec:refinement}

The contract produced by frame-constrained synthesis already contains the
projected frame conditions, but it may still fail the frame-error-free
checks. These failures can arise because the surrounding contract envelope
is incomplete: the candidate may lack validity preconditions needed to
justify an accessed region, loop annotations needed to prove an assigned
slice, or auxiliary assertions needed to expose a dependence relation to
the verifier. AutoFraming therefore performs verifier-guided refinement
using feedback from Frama-C/WP and Eva.

The refinement loop is frame-constrained. The LLM may revise the contract
to make the projected frame conditions checkable, but it may not delete,
weaken, or replace them with degenerate frame clauses.

In each round, the LLM reads the function body, the current contract, the
projected frame skeleton, and the verifier feedback, and proposes a
revised contract. A revised candidate $C_f^{(k+1)}$ is accepted only when
the following conditions hold.

First, the frame structure must be preserved. Every
\texttt{\textbackslash separated} precondition and every
\texttt{assigns}\,\ldots\,\texttt{\textbackslash from} clause projected
from $\Phi_f$ must remain present. The LLM may strengthen these clauses,
but it cannot delete them or replace them with a weaker frame.

Second, the candidate must not contain a degenerate frame. In particular,
AutoFraming rejects candidates with missing \texttt{assigns} clauses,
missing \texttt{loop assigns} clauses, or vacuous frames that hide
caller-visible writes.

Third, the candidate must make strict proof progress: it must close at
least one previously unresolved proof obligation without reopening any
obligation that was already discharged.

The first two conditions prevent the LLM from satisfying the verifier by
weakening the frame. The third keeps the repair loop making
monotone progress. The loop ends when Frama-C/WP and EVA discharge the relevant
frame conditions or when it reaches the maximum number of rounds $N$.

% ============================================================================
\section{Evaluation}
\label{sec:evaluation}

We evaluate AutoFraming with respect to three questions.

\textbf{RQ1 (Performance).} \emph{How effectively does AutoFraming synthesize frame specifications?} 

\textbf{RQ2 (Ablation).} \emph{ How do CPG-based frame analysis and frame-constrained projection contribute to the efficiency of AutoFraming?
}  

\textbf{RQ3 (Generality).} \emph{Can AutoFraming improve existing frameworks' capability of synthesizing frame specifications?}

\subsection{Datasets}
\label{sec:benchmarks}

We use the following existing datasets:

\paragraph{$\mathcal{D}_{\mathrm{a}}$: Real-world C code }
Nine real-world C programs from the Preguss public benchmark~\cite{wang2025},
across an embedded RTOS (Atomthreads), four
Contiki OS subsystems (ctimer, etimer, process, trickle), a TLS X.509
certificate parser (x509-submodule-2), an authenticated-encryption mode
(AES-CCM), and an embedded database (Antelope memb, Antelope relation). We include functions whose frame specifications are expressible in first-order ACSL without user-defined inductive predicates, yielding 84 benchmark functions. Functions requiring such predicates are excluded because their verification rests on the quality of a hand-written predicate, which is orthogonal to the frame synthesis we study in this paper. The same inclusion criterion is applied to all compared configurations. For transparency we also report the excluded set, where Preguss's error-free frame rate is only 3.3\% of the predicate-dependent functions, confirming that these cases are omitted because they need inductive-predicate reasoning outside our target problem, not because their exclusion advantages AutoFraming.

\paragraph{$\mathcal{D}_{\mathrm{b}}$: Frama-C Problems}
Fifty-one small C programs from the Frama-C tutorial suite widely used by formal verification works, organised into eight categories including arrays\_and\_loops, general\_wp\_problems,
immutable\_arrays, loops, miscellaneous,
more\_arrays, mutable\_arrays, pointers.

\subsection{Experimental Setup}
\label{sec:setup}
For each function we measure two frame-quality metrics.
\emph{Frame warning}: a warning emitted by the Frama-C WP plugin indicating
that the contract has no \texttt{assigns} clause or that the clause is
incomplete.
\emph{Frame fail}: reported by the Frama-C WP plugin together with the Eva
plugin; a function is frame-failing if it fails at least one
$\texttt{typed\_<f>\_assigns\_*}$ or
$\texttt{typed\_<f>\_loop\_assigns\_*}$ goal during WP discharge, or if it
fails the Eva dependence-set check.
We classify a function as frame-error-free only when all three checks
pass: (1) Frama-C emits no missing/incomplete assigns warning, (2) WP
discharges every generated assigns and loop-assigns proof obligation
within the timeout, and (3) Eva reports no dependency set mismatch for
the write-set locations. Timeouts on frame conditions are counted as
frame failures. We do not count a function as error-free merely because
the contract is syntactically accepted.
To confirm that strengthening a frame does not weaken the functional
specification, we also report the proof ratio (goals proved / total goals)
from the WP results.

To show that our results are not specific to a single model, we instantiate
the pipeline identically with two different models, Gemini-2.5-Pro and
GPT-5.2.
Verification uses Z3 and Alt-Ergo with a 30-second per-goal timeout.

\subsection{RQ1: How effectively does AutoFraming synthesize frame specifications?}
\label{sec:rq1}

RQ1 evaluates whether AutoFraming can synthesize frame specifications
that are accepted by frame specific checks. We instantiate the same
pipeline with two LLMs, Gemini-2.5-Pro and GPT-5.2, using the same configurations. We denote AutoFraming as AF in the result tables. Accordingly,
AF-GPT and AF-Gemini refer to AutoFraming instantiated with GPT-5.2 and
Gemini-2.5-Pro, respectively. We
compare against the strongest available ACSL specification generators on
our benchmarks: Preguss on both benchmarks and AutoSpec on the academic
benchmark. All methods are evaluated using the same frame-error-free
criterion.

Tables~II and~III report the number of functions for which each method
synthesizes a frame-error-free specification. On the real-world benchmark
$D_{a}$, AutoFraming substantially improves over Preguss: AF-Gemini
succeeds on 78 of 84 functions (92.9\%) and AF-GPT succeeds on 77 of 84
functions (91.7\%), compared with 28 of 84 functions (33.3\%) for
Preguss. On the academic benchmark $D_{b}$, AF-Gemini succeeds on 49
of 51 functions (96.1\%) and AF-GPT succeeds on 48 of 51 functions
(94.1\%), compared with 41 of 51 functions (80.4\%) for Preguss and 23
of 51 functions (45.1\%) for AutoSpec.

These results show that frame synthesis exposes failures that are not
captured by overall proof success alone. Preguss and AutoSpec can produce
contracts that are useful for functional verification, but their frame
clauses are often missing, incomplete, or insufficiently precise for the
frame-specific checks used in this study. AutoFraming improves because it
targets the frame conditions directly: region separation and data
dependence for write-set locations are synthesized before the LLM
synthesizes the rest of the contract.

Table~IV further decomposes each method's failures into frame warnings and
frame proof failures. The breakdown shows that the main weakness of the
baselines is frame incompleteness. On $D_{a}$, Preguss emits 51 frame
warnings and has 5 frame failures, leaving only 28 functions
frame-error-free. AutoFraming sharply reduces these warnings: AF-GPT has
2 warnings and 5 frame failures, while AF-Gemini has 4 warnings and 2
frame failures. On $D_{b}$, both AutoFraming variants leave only one
frame warning, compared with 10 for Preguss and 28 for AutoSpec.

The few functions AutoFraming leaves unverified falls into two categories. The first is a limit of the \emph{verifier}. For example, \texttt{memb\_alloc} writes
\texttt{m->count[0..m->num-1]}, an extent whose bound \texttt{m->num} is a
mutable struct field; the synthesized frame names exactly this region and is
correct, but the verifier cannot decide the completeness of a range whose bound is a
runtime value. 

The second failure type occurs when the LLM overrides the projected
frame skeleton with its own reading of the code during contract synthesis.
Although AutoFraming is designed to preserve the projected frame, the
final candidate still passes through an LLM generation step. In a small
number of cases, this generation changes the structure of the frame in a
way that remains syntactically valid but becomes harder for WP to
discharge. For example, \texttt{set\_iv}, our analysis gives its
write set as a single contiguous region, \texttt{assigns iv[0..15]}, and
AF-Gemini keeps it: WP discharges the frame at once. AF-GPT, handed the very
same input, re-derives the frame to mirror the four write statements in the
body---\texttt{iv[0]}, the thirteen bytes copied into \texttt{iv[1..13]} by
\texttt{memcpy}, \texttt{iv[14]}, and \texttt{iv[15]}, and emits them as four
disjoint slices. This
forces WP to re-establish the \texttt{memcpy}'s separation against four separate
regions, which causes the timeout. That is a limitation of our approach because our approach still depends on LLM generation.

\textbf{Answer to RQ1.} AutoFraming synthesizes frame-error-free
specifications for more than 90\% of functions on both benchmarks and
substantially outperforms existing ACSL generation frameworks under
frame-specific checks. The improvement is consistent across two different
LLMs, suggesting that the gain comes from the frame-synthesis design
rather than from a single model choice.

\begin{table}[t]
\caption{Error-free frame counts per program on the real-world benchmark $\mathcal{D}_{\mathrm{a}}$.AF-GPT and AF-Gemini denote AutoFraming instantiated with GPT-5.2
and Gemini-2.5-Pro, respectively.}
\label{tab:rq1-real}
\centering
\footnotesize
\setlength{\tabcolsep}{5pt}
\begin{tabular}{@{}lrrrrr@{}}
\toprule
Program & \#Fns & LOC & Preguss & AF-GPT & \textbf{AF-Gemini} \\
\midrule
AES-CCM             & 10 &  610 &  2 & 6   & \textbf{8}  \\
Antelope memb       &  1 &   55 &  0 & 1   & \textbf{0}  \\
Antelope relation   &  1 &   42 &  0 & 1  & \textbf{1}  \\
Atomthreads         & 10 &  830 &  1 & 10  & \textbf{9}  \\
Contiki ctimer      & 10 &  315 &  0 & 10  & \textbf{10} \\
Contiki etimer      & 10 &  195 &  8 & 10  & \textbf{10} \\
Contiki process     & 11 &  420 &  4 & 10   & \textbf{11} \\
Contiki trickle     & 11 &  395 &  3 & 11  & \textbf{11} \\
X.509 submod-2      & 20 & 1480 & 10 &18  & \textbf{18} \\
\midrule
\textbf{TOTAL}      & \textbf{84} & \textbf{4342}
                    & \textbf{28} & \textbf{77} & \textbf{78} \\
$\quad$frame-error-free rate
                    & & & 33.3\% & 91.7 \% & \textbf{92.9\%} \\
\bottomrule
\end{tabular}
\end{table}
 
\begin{table}[t]
\caption{Error-free frame counts per category on the academic benchmark $\mathcal{D}_{\mathrm{b}}$. AF-GPT and AF-Gemini denote AutoFraming instantiated with GPT-5.2
and Gemini-2.5-Pro, respectively.}
\label{tab:rq1-acad}
\centering
\footnotesize
\setlength{\tabcolsep}{4pt}
\begin{tabular}{@{}lrrrrr@{}}
\toprule
Category & \#Fns & AutoSpec & Preguss & AF-GPT & \textbf{AF-Gemini} \\
\midrule
arrays\_and\_loops    &  5 & 3 &  4 &5   & \textbf{5}  \\
general\_wp\_problems & 12 & 10 & 10 & 12 & \textbf{12} \\
immutable\_arrays     &  8 & 0 &  8 & 7  & \textbf{7}  \\
loops                 &  8 & 1 &  5 & 8  & \textbf{7}  \\
miscellaneous         &  5 & 2 &  4 & 4  & \textbf{5}  \\
more\_arrays          &  3 & 0 &  2 &3   & \textbf{3}  \\
mutable\_arrays       &  2 & 0 &  0 &  2 & \textbf{2}  \\
pointers              &  8 & 7 &  8 &7   & \textbf{8}  \\
\midrule
\textbf{TOTAL}        & \textbf{51}
                      & \textbf{23} & \textbf{41} & \textbf{48} & \textbf{49} \\
$\quad$frame-error-free rate
                      & & 45.1\% & 80.4\% & 94.1\% & \textbf{96.1\%} \\
\bottomrule
\end{tabular}
\end{table}
 
% \paragraph{Where the gains come from.}
% Table~\ref{tab:rq1-defects} drills into breakdown of frame spec warning and frame spec failure. On
% $\mathcal{D}_{\mathrm{a}}$, Preguss's $56$ residual defects are
% dominated by warnings ($51/56$, $91\%$): the tool emits no
% \texttt{assigns} clause or the assigns clause is incomplete. LLM-only shifts
% the mix: of its $32$ defects, $26$ still trigger warnings but $6$ are
% function-level \texttt{assigns} failures. Ours-synthesis closes both columns
% simultaneously, reducing warnings from $26$ to $4$ and failures from
% $6$ to $2$. On $\mathcal{D}_{\mathrm{b}}$, static generators
% (AutoSpec, Preguss) account for all their defects through warnings;
% LLM-only divides its $12$ defects evenly between warning and
% assigns-fail. Ours-synthesis collapses both columns on
% $\mathcal{D}_{\mathrm{b}}$ to a single warning and a single
% \texttt{assigns} failure, with the remaining two samples discussed in
% \emph{Remaining failures} below.
 
% --- Per-bucket breakdown table -----------------------------------------
\begin{table}[h]
\caption{Frame-outcome breakdown per method---frame warnings, frame failures, and error-free---with each row summing to \#Fns. AF-GPT and AF-Gemini denote AutoFraming instantiated with GPT-5.2
and Gemini-2.5-Pro, respectively. }
\label{tab:rq1-defects}
\centering
\footnotesize
\setlength{\tabcolsep}{5pt}
\begin{tabular}{@{}l l r r r@{}}
\toprule
                  &               & Frame                   & Frame      &   Frame           \\
                 
Benchmark         & Method        & Warning            & Fail           & Error-free        \\
\midrule
\multirow{3}{*}{$\mathcal{D}_{\mathrm{a}}$ (84)}
  & Preguss        & 51                 & 5                      & 28           \\
  & AF-GPT       & 2                 & 5                      &   77         \\
  & \textbf{AF-Gemini}  & \textbf{4}         & \textbf{2}             & \textbf{78}  \\
\midrule
\multirow{4}{*}{$\mathcal{D}_{\mathrm{b}}$ (51)}
  & AutoSpec       & 28                 & 0                      & 23           \\
  & Preguss        & 10                 & 0                      & 41           \\
  & AF-GPT       & 1                  &   2                    & 48           \\
  & \textbf{AF-Gemini}  & \textbf{1}         & \textbf{1}             & \textbf{49}  \\
\bottomrule
\end{tabular}
\end{table}

\subsection{RQ2: How do CPG-derived frame information and frame-constrained projection contribute to the efficiency of AutoFraming?
}
\label{sec:rq2}

% 1, report the numbers and answer the research question we raised. 2, show the gap categories between full and ablation variant to show all the rules really matters.

To isolate the source of AutoFraming's frame quality, we compare the full
pipeline with three reduced variants. The first variant removes
CPG-derived frame information but keeps frame-constrained projection and
frame-constrained refinement. The second keeps the CPG information but
removes frame-constrained projection. The third removes both CPG information
and frame-constrained projection, leaving only LLM-based refinement.

The full AutoFraming pipeline synthesizes frame-error-free specifications
for 78 of 84 functions. Removing the frame-constrained projection while
keeping the CPG information and the full frame-constrained refinement loop
reduces the result to 54 of 84 functions, a loss of 24 functions. This
drop shows that CPG information alone is not sufficient: without the
projection step, the LLM must still infer how the extracted information
should be translated into ACSL frame clauses.

The other reduced variants support the same conclusion. The configuration
without frame-constrained projection reaches 54 functions, which is close
to the refinement-only configuration with 53 functions. Removing the
CPG-based frame information reduces the result further to 48 functions.
Thus, neither CPG information alone nor frame-constrained refinement alone
explains AutoFraming's improvement. The main gain comes from the
frame-constrained projection.

The two frame constraints govern where the variants fail. Most failures fall
under the data dependence projection for write sets, in two forms. Some are
\emph{data-dependence} errors: the model names the right locations but omits
the \texttt{\textbackslash from} dependence (e.g.\ \texttt{assigns *dest;}
instead of \texttt{assigns *dest \textbackslash from *src;}), or it declares
an unsound dependence, as when \texttt{set\_iv} gives 
\texttt{\textbackslash from nonce[0..12]} that cannot be discharged through the
library \texttt{memcpy}. Others are \emph{write-set} errors, where the extent
of the write set is wrong rather than merely its name:
\texttt{aes\_128\_set\_padded\_key} and \texttt{set\_key} declare
\texttt{assigns round\_keys[0..10]} but miss the inner array dimension---the
sound frame is \texttt{round\_keys[0..10][0..15]}; \texttt{atomCurrentContext},
\texttt{list\_head}, and \texttt{arg\_alloc} omit the
\texttt{\textbackslash result} location entirely; and \texttt{\_parse\_arc}
declares \texttt{assigns \textbackslash nothing} for a function that writes two
outputs. The remaining failures fall under \emph{region separation}. For example,
\texttt{parse\_PolicyInformation} and \texttt{parse\_policyQualifierInfo}
declare the correct write set but hedge their precondition, weakening the
\texttt{\textbackslash separated} guarantee until the frame condition can no
longer be discharged.
 
\noindent\textit{Answer to RQ2.}
The ablation shows that the frame-constrained projection is the central
component of AutoFraming. Removing this projection reduces the
frame-error-free count from $78/84$ to $54/84$, a loss of 24 functions.
This indicates that CPG information and frame-constrained refinement are
not sufficient by themselves; the extracted information must be projected
into explicit ACSL frame conditions before the LLM assembles the final
contract. The refinement-only configuration reaches a similar result
($53/84$), and removing CPG-based frame information lowers the result to
$48/84$. Thus, AutoFraming's improvement comes primarily from the
frame-constrained projection.

\begin{table}[t]
\caption{RQ2: Ablation on $\mathcal{D}_{\mathrm{a}}$ error-free frame counts for AutoFraming and three reduced variants. }
\label{tab:ablation}
\centering
\footnotesize
\setlength{\tabcolsep}{8pt}
\begin{tabular}{@{}lrr@{}}
\toprule
Configuration & \#Clean & Rate \\
\midrule
LLM Refinement Only           & 53/84 & 63.1\% \\
Without CPG derived info           & 48/84 & 57.1\% \\
Without frame-constrained projection           & 54/84 & 64.3\% \\
\textbf{AutoFraming} & \textbf{78/84} & \textbf{92.9\%} \\
\bottomrule
\end{tabular}
\end{table} 

% \FloatBarrier
\subsection{RQ3: Can AutoFraming improve  existing frameworks as a frame synthesis component?}
\label{sec:rq3}
 % --- RQ2 wide table declared here (BEFORE the FloatBarrier) so it
%     flushes with RQ1's queued floats and lands at the top of an
%     appropriate page, instead of being deferred into RQ2 prose
%     and trapped at the bottom with blank space above. -----------------
\begin{table}[!t]
\caption{RQ3: error-free frame counts and goal-prove rates when
AutoFraming augments existing generators. ``+LLMRefine'' denotes direct
LLM refinement without AutoFraming's frame-constrained projection, and
``+AutoFraming'' denotes refinement with AutoFraming.}
\label{tab:rq3}
\centering
\footnotesize
\setlength{\tabcolsep}{3pt}
\begin{tabular}{@{}llrrr@{}}
\toprule
Benchmark & Upstream & \#Fns & Error-free frame & Goal-prove \\
\midrule
\multirow{3}{*}{$\mathcal{D}_{\mathrm{a}}$}
  & Preguss          & 84 & 28 (33.3\%) & 94.0\% \\
  & Preguss+LLMRefine       & 84 & 52 (61.9\%) & 93.2\% \\
  & \textbf{Preguss+AutoFraming}
                     & 84 & \textbf{80 (95.2\%)} & \textbf{96.8\%} \\
\midrule
\multirow{6}{*}{$\mathcal{D}_{\mathrm{b}}$}
  & AutoSpec                  & 51 & 23 (45.1\%) & 86.4\% \\
  & AutoSpec+LLMRefine
                              & 51 & 36 (70.6\%) & 89.4\% \\
  & \textbf{ AutoSpec+AutoFraming}
                              & 51 & \textbf{50 (98.0\%)} & \textbf{97.9\%} \\
  & Preguss                   & 51 & 41 (80.4\%) & 96.0\% \\
  & Preguss+LLMRefine & 51 & 39 (76.5\%) & 93.7\% \\
  & \textbf{Preguss+AutoFraming}
                              & 51 & \textbf{50 (98.0\%)} & \textbf{97.5\%} \\
\bottomrule
\end{tabular}
\end{table}
 
RQ3 evaluates whether AutoFraming can be used as a modular
frame-synthesis component for existing specification generators. Rather
than replacing upstream tools such as Preguss and AutoSpec, AutoFraming
augments their generated contracts by strengthening the frame conditions. Table~VI reports the
frame-error-free counts and goal-prove rates when AutoFraming augments
Preguss and AutoSpec. Three observations show that AutoFraming improves the frame synthesis of these tools without degenerating their functional specification.

First, the total number of verification goals increases after applying
AutoFraming: from 1378 to 1687 when comparing Preguss with
Preguss+AutoFraming on $\mathcal{D}_{\mathrm{a}}$, from 663 to 681 when
comparing AutoSpec with AutoSpec+AutoFraming on
$\mathcal{D}_{\mathrm{b}}$, and from 676 to 786 when comparing Preguss
with Preguss+AutoFraming on $\mathcal{D}_{\mathrm{b}}$. The added goals
come from the \texttt{\textbackslash separated} preconditions and \texttt{\textbackslash from}
dependencies synthesized by AutoFraming. A weakened contract would tend
to remove obligations rather than introduce additional ones.

Second, the goal-prove rate also increases despite the larger number of
goals. On $\mathcal{D}_{\mathrm{a}}$, Preguss+AutoFraming raises the
goal-prove rate from $94.0\%$ to $96.8\%$. On
$\mathcal{D}_{\mathrm{b}}$, AutoSpec+AutoFraming raises the rate from
$86.4\%$ to $97.9\%$, and Preguss+AutoFraming raises it from $96.0\%$ to
$97.5\%$. Thus, AutoFraming discharges more proof obligations in absolute
terms while also increasing the proved-to-total ratio.

Third, AutoFraming outperforms a direct LLM-refinement baseline. On
$\mathcal{D}_{\mathrm{b}}$, AutoSpec+LLMRefine reaches $36/51$
frame-error-free functions ($70.6\%$), while AutoSpec+AutoFraming reaches
$50/51$ ($98.0\%$), a $27.4$ percentage-point improvement. Similarly,
Preguss+LLMRefine reaches $39/51$ ($76.5\%$), while
Preguss+AutoFraming reaches $50/51$ ($98.0\%$), a $21.5$ percentage-point
improvement. This shows that the gain does not come from the LLM only refinement; it comes from AutoFraming's
frame-constrained projection and refinement.

\noindent\textit{Answer to RQ3.}
AutoFraming generalizes as a frame-specification refinement layer for
existing ACSL generators. With Preguss and AutoSpec as upstream tools, it
raises the frame-error-free rate to $95.2\%$--$98.0\%$ while also
increasing the goal-prove rate and the absolute number of proved goals.
This indicates that AutoFraming strengthens the frame specification
rather than improving the frame score by weakening the surrounding
contract. We do not claim that the goal-prove rate establishes semantic
completeness of the functional clauses; rather, it checks that the added
frame clauses remain compatible with the evaluated verifier.
 
\FloatBarrier

% \begin{table}[t]
% \caption{RQ3: Detection of injected frame-violating mutations on the
% 33-function silent-FN cohort (functions where all three tools
% reported full verification before mutation). Higher detection rate
% means more bugs caught; the cohort represents the worst case for the
% baselines because their unmutated contracts already verified.}
% \label{tab:rq3}
% \centering
% \footnotesize
% \setlength{\tabcolsep}{8pt}
% \begin{tabular}{@{}lrrr@{}}
% \toprule
% Tool          & \#Fns & Caught & Detection rate \\
% \midrule
% Preguss       & 33 &  9 & 27.3\% \\
% LLM-refine     & 33 & 28 & 84.8\% \\
% \textbf{Ours-refine}
%               & 33 & \textbf{30} & \textbf{90.9\%} \\
% \bottomrule
% \end{tabular}
% \end{table}

\FloatBarrier

\section{Related Work}
\label{sec:related}

\subsection{Program-Analysis-Based Specification Inference}

Early work on specification inference mainly relies on program analysis to derive likely contracts, invariants, or behavioral properties from source code. These approaches can be broadly characterized by whether they reason from observed executions or from static program semantics.

Dynamic inference techniques generate candidate specifications from execution traces. Daikon~\cite{Ernst2007}, for example, instruments programs and infers likely invariants from recurring value patterns observed at program points such as function entries and exits. CLIPER~\cite{lo2007efficient} similarly mines temporal behavioral patterns from traces. Although such techniques are useful for discovering common runtime properties, their results depend heavily on input coverage and may overfit to observed executions. To improve generalization, later systems combine concrete traces with symbolic reasoning. DySy~\cite{Csallner2008} integrates dynamic execution and symbolic execution, using counterexamples to refine inferred invariants, while SymInfer~\cite{Nguyen2022} combines concrete traces with symbolic states to produce more general specifications. SPEEDY~\cite{Cok_2014} further connects runtime monitoring with Frama-C static analyzers to suggest specifications during development.

Static approaches infer specifications without executing the program, often aiming to cover all feasible paths. Kindspec~\cite{alpuente2020abstract,alpuente2023automated} applies symbolic execution to KernelC programs and synthesizes abstract preconditions and postconditions as logical contracts. TriCera~\cite{amilon2021automated} translates C programs into constrained Horn clauses and uses SMT solving to infer ACSL function contracts. AutoDeduct~\cite{Amilon2025} also automates contract generation through symbolic execution and Horn clause reasoning. In contrast to these automated techniques, Araujo et al.~\cite{dordowsky2015experimental} demonstrate a manual specification and verification process by translating avionics requirements into ACSL annotations and checking them with Frama-C. These works show the value of program analysis for contract inference, but they generally focus on functional behavior rather than frame specifications.

\subsection{LLM-Assisted Formal Specification Generation}

Recent work has explored large language models as a flexible mechanism for generating formal specifications from code, natural language requirements, examples, or verifier feedback. For C programs, George et al.~\cite{George2025Seeking} prompt GPT-4 to synthesize ACSL contracts directly from source code, showing that LLMs can often produce specifications that capture intended behavior rather than merely restating implementation details. In follow-up work, George et al.~\cite{George2025Specify} enrich the prompt with outputs from Frama-C Eva~\cite{frama-c-eva} and PathCrawler~\cite{frama-c-pathcrawler}, demonstrating that static-analysis warnings and test-based traces can guide the model toward more useful contracts.

Several systems combine LLM generation with formal-verification feedback. AutoSpec~\cite{Cheng2024} constructs program information such as call graphs and iteratively refines LLM-generated specifications using verifier feedback. Jan{\ss}en et al.~\cite{Christian2023} evaluate ChatGPT on small C programs and show that generated loop invariants can help Frama-C and CPAchecker complete proofs, while also identifying common failure modes such as weak or invalid invariants. Guangyuan et al.~\cite{Guangyuan2024} use GPT-3.5 to propose loop invariants and rely on CBMC counterexamples to guide refinement. ACInv~\cite{liu2024enhancing} extracts loop context through static analysis, prompts an LLM to generate invariants, and uses an LLM-based evaluator to iteratively strengthen, weaken, or reject candidates. NL2Contract\cite{richter2025beyond} translates informal natural-language hints into formal functional contracts with both preconditions and postconditions for automatic software verification. Preguss~\cite{wang2025} improves the scalability of LLM-assisted specification synthesis by using potential runtime-error assertions to decompose large C verification tasks into prioritized units. These studies suggest that LLMs are most effective when combined with symbolic information and verifier feedback, but their primary focus remains on functional contracts and loop invariants.

LLM-based specification generation has also been studied beyond C and ACSL. For Java, Teuber and Beckert~\cite{Teuber2025} use GPT to complete JML annotations and repair invalid contracts with feedback from the KeY verifier, while Greiner et al.~\cite{greiner2024automated} train CodeT5-based models to generate JML preconditions and postconditions from a large corpus of examples. SpecGen~\cite{Lezhi2025} adopts a two-stage process that first prompts an LLM to draft specifications and then uses mutation and heuristic selection to repair them. For Rust, SAFE~\cite{chen2024automated} generates Verus specifications from function signatures and documentation, followed by iterative self-refinement. ClassInvGen~\cite{Chuyue2025} targets C++ class invariants and validates candidates through execution. In the smart-contract domain, PropertyGPT~\cite{Ye2025} retrieves examples from Certora reports to guide GPT-4 in generating Solidity properties, which are then checked and refined by custom provers. VeCoGen~\cite{sevenhuijsen2025vecogen} studies the related problem of generating C programs from ACSL specifications, natural-language descriptions, and tests.

% \subsection{Motivation: Frame Specifications Remain Underexplored}

% Despite recent progress, existing specification-generation methods mainly emphasize functional properties, such as preconditions, postconditions, loop invariants, and assertion-like properties. Frame specifications, which describe memory side effects, data dependencies, and separation constraints, receive comparatively less attention. In ACSL-based modular verification, however, incomplete frame specifications can prevent tools from precisely reasoning about which locations may be modified and which values may influence outputs. This can lead to unstable caller-side proofs, imprecise contracts, or specifications that appear syntactically valid but are insufficient for reusable verification.

% This gap motivates our work. Instead of treating frame clauses as secondary annotations, we explicitly target ACSL frame specification synthesis. Our approach uses program-structure and data-flow information extracted from code property graphs to guide LLM-based generation and refinement, with the goal of producing more complete frame specifications suitable for modular verification.

\section{Threats to Validity}
\label{sec:threats}

\paragraph{Internal validity}
Internal threats mainly concern LLM behavior and the implementation of our analysis rules. Since LLMs are trained on large code and specification corpora, benchmark or similar examples may have appeared during pretraining and could inflate performance. To mitigate this concern, our ablation with only LLM-based refinement, without CPG-derived information or frame-constrained projection, performs substantially worse, suggesting that the gains mainly come from the proposed frame analysis and rules. Another threat is that our CPG-based rules are derived from frame-relevant program information and expert knowledge, but may still miss uncommon C patterns such as complex pointer manipulation or unusual data dependencies. Missing rules or implementation errors may lead to incomplete frame specifications or unresolved verification goals.

\paragraph{Construct validity}
Our main evaluation metric is error-free frame status, which is determined using Frama-C/WP diagnostics and Frama-C's dependency analysis. This metric provides practical information that the generated frame specifications are accepted by the verification toolchain, but it is not a complete semantic oracle for all possible client contexts. For example, a generated frame specification may verify successfully while imposing preconditions that are stronger than necessary and may exclude some valid calls. Similarly, the goal-prove rate measures verifier discharge rather than specification minimality or user intent. Therefore, our results should be interpreted as verifier-accepted frame correctness under the evaluated toolchain.

\paragraph{External validity}
Our experiments are conducted on 84 functions from real-world C systems and 51 academic verification examples. Although these benchmarks cover common frame-relevant patterns such as pointers, arrays, loops, and data dependencies, they may not represent all C programs. The results may not generalize to software with complex low-level features, such as unrestricted casts, concurrency, etc. In addition, our approach is designed for ACSL and Frama-C/WP; applying it to other specification languages, verification tools, or programming languages may require adapting the frame rules and verification workflow. Finally, LLM behavior can vary across model versions, prompts, and decoding settings, so future work should evaluate additional models and release prompts, generated specifications, tool versions, and verification scripts to support replication.

\section{Conclusion}
\label{sec:conclusion}
We have presented AutoFraming for synthesizing ACSL frame specifications for C programs. It extracts frame-related information and projects the information into a deterministic frame skeleton, and uses the LLM only to complete and repair the surrounding contract under verifier feedback. Our experimental results show that grounding LLM-based specification synthesis in explicit program structure and frame-specific constraints leads to more complete and reusable frame specifications than existing generators. Future work will extend AutoFraming to richer heap abstractions, recursive data structures, inter-procedural library summaries, and additional specification languages and verifiers.

% Forces a page break and prints all pending figures/tables
\clearpage
% ============================================================================
% Bibliography (IEEEtran style)
% ============================================================================
\bibliographystyle{IEEEtran}
\bibliography{references}
\end{document}